\pdfoutput=1

\documentclass[10pt, a4paper]{article}

\usepackage{lrec-coling2024}

\usepackage{times}
\usepackage{latexsym}

\usepackage{latexsym}
\usepackage{enumerate}
\usepackage{numprint}
\usepackage{hyperref}
\usepackage{paralist}
\usepackage{caption}

\usepackage[T1]{fontenc}

\usepackage[utf8]{inputenc}

\usepackage{microtype}

\usepackage{inconsolata}

\usepackage{amsmath,amssymb}
\usepackage{tikz}
\usepackage{pgfplots}
\usepackage{pgfmath}
\usetikzlibrary{plotmarks,shapes,pgfplots.dateplot}
\usetikzlibrary{math,calc,trees,positioning,arrows,chains,shapes.geometric,
decorations.pathreplacing,decorations.pathmorphing,shapes,
matrix,shapes.symbols}
\usetikzlibrary{3d,decorations.text,shapes.arrows,fit,backgrounds,quotes,arrows.meta}
\usepackage{ifthen}
\usepackage{multirow,multicol}
\usepackage{booktabs}
\usepackage{float}
\usepackage{longtable}
\usepackage{graphicx}
\usepackage{caption}
\usepackage{subcaption}
\usepackage{url}

\colorlet{color01}{red!10!white}
\colorlet{color02}{orange!10!white}
\colorlet{color03}{green!10!white}
\colorlet{color04}{blue!10!white}
\colorlet{color05}{magenta!20!white}
\colorlet{color06}{black!15!white}
\colorlet{color07}{yellow!10!white}
\colorlet{color08}{white}
\colorlet{color1}{red!80!black}
\colorlet{color1a}{red!50!white}
\colorlet{color2}{orange!80!black}
\colorlet{color2a}{orange!50!white}
\colorlet{color3}{green!50!black}
\colorlet{color3a}{green!90!black}
\colorlet{color4}{blue!80!black}
\colorlet{color4a}{blue!50!white}
\colorlet{color5}{magenta!80!black}
\colorlet{color5a}{magenta!50!white}
\colorlet{color6}{black!80!black}
\colorlet{color6a}{black!50!white}
\colorlet{color7}{white!30!black}
\colorlet{color7a}{black!50!white}
\colorlet{color8}{yellow}
\colorlet{color8a}{black!50!yellow}

\tikzset{every picture/.style={semithick},every path/.style={thick,rounded corners,->}}
\tikzset{
token/.style={rectangle,rounded corners,draw=black,thick,inner sep=4pt,outer sep=0,minimum width=2em,fill=color03,text centered, minimum height=1.5em,font=\small\sffamily},
tokenperm/.style={token,fill=color04},
tokenres/.style={token,fill=color02},
numperm/.style={token,fill=color08},
ne/.style={ draw=none, fill=none, font=\footnotesize\sffamily,  minimum height=0em, text centered},
ncirc/.style={ circle, draw=black, thin, fill=none, font=\footnotesize\sffamily,  minimum height=2em, inner sep=0, text centered},
nd/.style={ font=\sffamily,  text centered},
nodecomp/.style={ rectangle,  rounded corners,  draw=black, thick, text width=2em,  font=\footnotesize\sffamily, minimum height=1.3em,  text centered},
nodevar/.style={ nodecomp, fill=green!10,
},
diablo/.style={ rectangle,  rounded corners,  draw=black, thick, text width=10em,  font=\footnotesize\sffamily, minimum height=1em,  text centered},
branch/.style ={circle,inner sep=0pt,minimum size=1.5mm,fill=black,draw=black},
diablo2/.style={ rectangle,  rounded corners,  fill=red!10, draw=black!80,thick, text width=3em,  font=\footnotesize\sffamily, minimum height=2.7em,  text centered},
diafeature/.style={ rectangle, rounded corners=2pt,  fill=green!10, draw=black!80,thick, text width=1em,  font=\footnotesize\sffamily, minimum height=6em,  text centered},
diafeatnarr/.style={ rectangle, rounded corners=2pt,  fill=green!10, draw=black!80,thick, text width=.5em,  font=\footnotesize\sffamily, minimum height=6em,  text centered},
dialoss/.style={ diablo2, fill=green!10,
},
rotnode/.style={ anchor=center, rotate=90, font=\footnotesize\sffamily
},
diaext/.style={ diablo2,fill=yellow!40, },
diablo3/.style={rectangle, rounded corners, fill=blue!10, draw=blue!40,thick, text width=3.5em,  font=\footnotesize\sffamily\bfseries, text=blue, minimum height=1.5em, text centered},
line/.style={draw=red,rounded corners,thick, ->, decoration={markings,mark=at position 1 with {\arrow[scale=4,>=stealth]{>}}},postaction={decorate}},
element/.style={ tape, top color=white, bottom color=blue!50!black!60!, minimum width=8em, draw=blue!40!black!90, very thick, text width=10em, minimum height=3.5em, text centered, on chain},
every join/.style={->,rounded corners,thick,shorten >=1pt},  decoration={brace},
lineblue/.style={    join,line width=.07cm,->,blue!20  }
}

\usepackage{algorithm2e}

\SetCommentSty{mycommfont}
\SetKwComment{Comment}{/* }{ */}
\RestyleAlgo{ruled}

\def\dataset{\ensuremath{\text{SearchBySnippet}}\xspace}
\def\model{\ensuremath{\text{SnippeR}}\xspace}
\def\modelb{\ensuremath{\textbf{SnippeR}}\xspace}

\begin{document}

\title{Searching by Code: a New SearchBySnippet Dataset and SnippeR Retrieval Model for Searching by Code Snippets}

\name{Ivan Sedykh$^1$, Dmitry Abulkhanov$^2$, Nikita Sorokin$^1$,\\
\large \textbf{Sergey Nikolenko$^{3,4}$, Valentin Malykh$^{1,4}$}}

\address{$^1$ Huawei Noah's Ark lab, $^2$ Independent Researcher, \\
$^3$ St. Petersburg Department of the Steklov Institute of Mathematics,\\
$^4$ Ivannikov Institute for System Programming \\
valentin.malykh@phystech.edu\\}

\abstract{
Code search is an important and well-studied task, but it usually means searching for code by a text query. We argue that using a code snippet (and possibly an error traceback) as a query while looking for bugfixing instructions and code samples is a natural use case not covered by prior art. Moreover, existing datasets use code comments rather than full-text descriptions as text, making them unsuitable for this use case. We present a new SearchBySnippet dataset implementing the search-by-code use case based on StackOverflow data; we show that on SearchBySnippet, existing architectures fall short of a simple BM25 baseline even after fine-tuning. We present a new single encoder model SnippeR that outperforms several strong baselines on SearchBySnippet with a result of 0.451 Recall@10; we propose the SearchBySnippet dataset and SnippeR as a new important benchmark for code search evaluation.
\\ \newline 
}

\maketitleabstract

\section{Introduction}

\begin{figure*}\centering
\includegraphics[width=\linewidth]{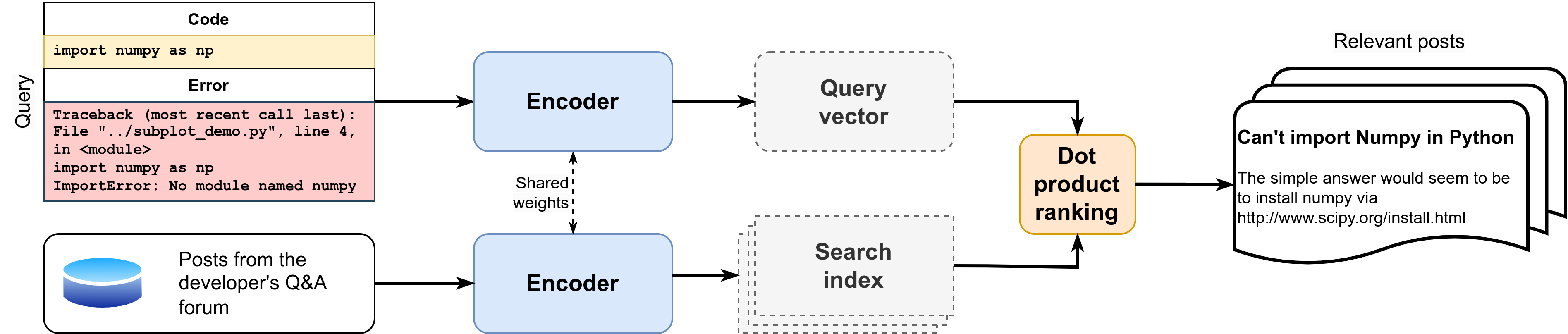}
\caption{Overview of the problem setting and system design.}
\label{fig:model-overview}
\end{figure*}

Increasing amounts of source code written every day lead to a plethora of possible issues, which almost inevitably
have already been solved and reported upon on forums such as \emph{StackOverflow}. A developer debugging an error has the relevant code snippet and error traceback produced by the compiler or interpreter, and
she wants to find out the reasons behind the error and ways to fix it.
This leads to the setting that we call ``search by snippet'': based on a code snippet and/or error traceback, find posts that might contain a solution.
To our surprise, this setting has been very scarcely considered in literature; e.g.,~\citet{ponzanelli2014mining} consider it in informally and just use
a commercial search engine. In this work, we propose an information retrieval setup where the query is a code snippet and/or traceback and documents are posts with text and possibly other code snippets (Fig.~\ref{fig:model-overview});
this setting can be automated and incorporated into IDEs.
Previous works on code search (see Section~\ref{sec:related}) usually matched the source code of a function and its comment, and
code search
also considers text queries; one can invert the problem but the text parts are usually short comments rather than full-text posts that could contain a solution.

In this work, we present a new dataset called \dataset that captures this problem setting based on \emph{StackOverflow} posts (in \emph{Python}). We have adapted several state of the art code search models as baselines, including CodeBERT,
GraphCodeBERT (GCB),
and SynCoBERT.
To our surprise, their performances on \dataset are very poor; even GCB specially trained on the CodeSearchNet dataset for this setting lost very significantly to the simple BM25 baseline.
Therefore, we have developed a new \model model that uses a GCB-based encoder for both queries and documents and incorporates a number of improvements so it outperforms BM25 on \dataset.
Still, absolute values of the results are not too high, and we believe that the problem setting embodied in \dataset opens up a new research direction that could lead to
better code understanding.

The primary contributions of this work include:
\begin{inparaenum}[(i)]
\item a novel problem setting for code search and a new \dataset dataset for training and evaluation in this setting;
\item the \model model that outperforms strong information retrieval baselines and can serve as a starting point for research in this new setting\footnote{We are going to release the \dataset and \model source code once the clearance is done.}.
\end{inparaenum}
Below, Section~\ref{sec:related} surveys related work, Section~\ref{sec:data} presents \dataset, Section~\ref{sec:model} introduces \model and its training procedure, Section~\ref{sec:eval} shows our experimental setup and results, and
Section~\ref{sec:conclusion} concludes the paper.

\section{Related Work}\label{sec:related}

\textbf{Datasets}.
\citet{husain_codesearchnet_2019} presented \emph{CodeSearchNet} (CSN), constructed from a \emph{GitHub} dump, with function bodies split into the code itself and a description. CSN contains 2M (code snippet, description) pairs in $6$ programming languages, including \emph{Python}.
\citet{hasan2021codesc}
combined CSN and other datasets
into a larger one (with \emph{Java} and \emph{Python} subsets of CSN),
getting 4M (code snippet, description) pairs. An even larger dataset had been constructed earlier by~\citet{gu2018deep}; their \emph{CODEnn-Train} \emph{Java}-based dataset has 18M pairs of methods and their one-sentence descriptions.
\emph{CodeXGLUE} by~\citet{lu_codexglue_2021} is a machine learning benchmark collection of datasets for code understanding and generation tasks, which includes
code in $10$ programming languages (and a modification of CSN). Another multi-task dataset was presented by~\citet{puri2021codenet}, with 14M code snippets in 5 programming languages.

\textbf{Code Search}.
Dense vector representations are often used for information retrieval (IR):
\citet{gu2018deep} used two RNNs to represent the code and textual descriptions, \citet{codebert} based CodeBERT
on language models, \citet{gotmare2021cascaded} used three Transformer-based models, two encoders and one classifier, to obtain a hierarchical representation of code and text.
Our model uses a single encoder for embedding both queries and documents and has no separate classifier.

\textbf{Language models for code}.
GraphCodeBERT~\cite{guo_graphcodebert_2021} uses data flow graphs during pretraining to solve masked language modeling, edge prediction, and node alignment tasks.
SynCoBERT~\cite{wang_syncobert_2021} uses multimodal contrastive learning to achieve better code representations and is pretrained on identifier prediction and abstract syntax tree (AST) edge prediction.

\section{\dataset Dataset}\label{sec:data}

\noindent
\textbf{Data Preprocessing}.
\dataset is constructed from a public \emph{StackOverflow} dump\footnote{\url{https://archive.org/details/stackexchange}}
with questions and answers from 2014 to 2021 and rich meta-information.
During submission, \emph{StackOverflow} users fill in several fields that appear in the dump structure (along with fields such as ``FavouriteCount''); the ``tags'' field allows to easily categorize questions. We limit our work to \emph{Python} due to its popularity.
For preprocessing, we take the ``\emph{text}'' and ``\emph{title}'' fields that contain the main text of a question (``text'' can have formatting markup)
and the $\langle\textit{code}\rangle$ tag for source code and/or system output and extract text from these tags. If it does not look like a traceback (e.g., does not have the ``\texttt{Error}'' keyword), we mark it as ``\emph{code}'' and extract in the ``\emph{code}'' field; if it does, we use the ``\emph{error}'' field. We also parse the error type from the traceback with regular expressions and put it into the ``\emph{keyword}'' field. If a question contains several $\langle\textit{code}\rangle$ tags, they are classified independently. We add a ``\emph{best\_answer}'' field for the answer accepted by the user and store the original text in the ``\emph{body}'' tag. Fig.~\ref{fig:model-overview} shows a sample preprocessed query on the left and a document on the right.

\begin{table}[!t]
\centering\setlength{\tabcolsep}{5pt}
\resizebox{\linewidth}{!}{
\begin{tabular}{l|r|r}
\midrule
\textbf{Constraint} & \textbf{Number} & \textbf{\%} \\
\midrule
\dataset, total & \numprint{948749} & 100.00 \\
$\qquad$with ``\emph{code}'' & \numprint{670561} & 70.68 \\
$\qquad$with ``\emph{error}'' & \numprint{478992} & 50.49 \\
$\qquad$with ``\emph{code}'' and ``\emph{error}'' & \numprint{342086} & 36.06\\
$\qquad$with ``\emph{code}'' or ``\emph{error}'' & \numprint{807467} & 85.10 \\
$\qquad$with ``\emph{best\_answer}'' & \numprint{337797} & 35.60 \\ \midrule
CodeSearchNet, total & \numprint{2070536} & 200.18 \\
$\qquad$Python only & \numprint{457461} & 48.22 \\ \midrule
NeuralCodeSearch, evaluation & \numprint{287} & $<$ 0.01 \\\bottomrule
\end{tabular} }

\caption{\dataset dataset statistics and comparison with other code search datasets.}
\label{tab:dataset1}
\end{table}

Table~\ref{tab:dataset1} shows the dataset statistics.
Only half of the questions have error tracebacks, and over 70\% contain code. Interestingly, questions where we have extracted both ``\emph{code}'' and ``\emph{error}'' fields cover only $1/3$ of the dataset, while the ones with ``\emph{code}'' \emph{or} ``\emph{error}'' cover $85\%$. Only $35\%$ of the questions have accepted answers. Table~\ref{tab:dataset2} shows the average sizes (in symbols) for extracted fields and compares them with
the ``\emph{body}'' field (percentages only count questions where the field(s) are present).

\begin{table}[!t]
\centering\setlength{\tabcolsep}{3pt}
\resizebox{\linewidth}{!}{
\begin{tabular}{l|r|r|r}
\midrule
\textbf{Subset} & \textbf{Total} & \textbf{Field} & \textbf{\% of} \\
& \textbf{size} & \textbf{size} & \textbf{total} \\
\midrule
\multicolumn{4}{c}{\textbf{\dataset}} \\\midrule
All posts & 1340.18 & 1340.18 & 100.00 \\
$\quad$with ``\emph{code}'' & 1630.62 & 576.26 & 35.34 \\
$\quad$with ``\emph{error}'' & 1981.50 & 681.04 & 34.37 \\
$\quad$with ``\emph{code}'' and ``\emph{error}'' & 2223.89 & 1366.80 & 61.46\\
$\quad$with ``\emph{best\_answer}'' & 1297.66 &  880.98 & 67.89\\
\midrule
\multicolumn{4}{c}{\textbf{CodeSearchNet} (\% of \dataset)} \\\midrule
``\emph{func\_code\_string}'' &  & 755.59 & 55.38 \\
$\quad$Python only &  & 1060.22 & 77.57 \\
{``\emph{func\_documentation\_string}''} &  & 209.87 & 23.82 \\
$\qquad$Python only &  & 297.59 & 33.78 \\\midrule
\multicolumn{4}{c}{\textbf{NeuralCodeSearch} (\% of \dataset)} \\\midrule
Answer &  & 159.75 & 11.59 \\
Question &  & 50.69 & 5.75\\
\bottomrule
\end{tabular} }

\caption{Average field sizes, in symbols; ``Field size'' shows the size of the field in ``Subset''.}
\label{tab:dataset2}
\end{table}

\textbf{Evaluation Set}.
Some questions in the \emph{StackOverflow} dump are marked as duplicates; usually post A is a duplicate of post B if \emph{StackOverflow} moderators have deemed the question in post A to be equivalent to the question in post B. We selected duplicated questions that contain an accepted answer (in post B) and a code snippet or traceback, getting $1369$ questions that we use for evaluation. We use a union of the ``\emph{code}'' and ``\emph{error}'' fields from post A as query and ``\emph{best\_answer}'' from post B as the ground truth document.

\textbf{Comparison to Other Datasets}.
We compare our dataset to CodeSearchNet~\cite{husain_codesearchnet_2019} also devoted to code search. It contains snippets in several programming languages, including \emph{Python}, in the form of functions paired with their descriptions. We also consider \emph{NeuralCodeSearch}~\cite{li2019neural} as the dataset with the most similar design; we only use its evaluation part of this dataset that contains \emph{StackOverflow} questions with code snippets cut out from the best answers.

Tables~\ref{tab:dataset1} and~\ref{tab:dataset2} compare \emph{CodeSearchNet} (CSN) and \emph{NeuralCodeSearch} (NCS) with \dataset. CSN is twice larger overall, but its \emph{Python} part is twice as \emph{small} as \dataset. NCS contains only 287 questions in its evaluation part, 3500x less than \dataset. Table~\ref{tab:dataset2} compares CSN and NCS with \dataset in terms of the average size of various fields (in symbols);
we assume that ``\emph{func\_code\_string}'' in CSN is a rough equivalent of the union of our ``\emph{code}'' and ``\emph{error}'', and ``\emph{func\_documentation\_string}'' corresponds to ``\emph{best\_answer}''. For NCS, the ``\emph{answer}'' and ``\emph{question}'' fields are inverted since ``\emph{best\_answer}'' in our case is a text field, while in NCS it is a code snippet. CSN and NCS parts of Table~\ref{tab:dataset2} show percentages of the corresponding (code and text) average field sizes in \dataset; while code-containing fields in CSN are only 20\% shorter, the text field is 3x to 4x times shorter than in \dataset. We believe that this could make retrieval on \dataset more difficult. In NCS, both entities are an order of magnitude shorter, leading to a much easier retrieval task.

\section{Model}\label{sec:model}

\textbf{Problem setting}.
In our IR task, the query is a code snippet and/or traceback from the ``\emph{code}'' and ``\emph{error}'' fields and documents are answers from the ``\emph{best\_answer}'' field. Given a collection of documents ${D}$ and a query $q$, the model has to rank documents so that the ground truth answer $d \in {D}$ is closer to the beginning of the list.
Following prior art on code search, we use a neural network encoder to obtain dense vector representations of queries and documents. Unlike~\citet{DPR}, and following~\citet{codebert} in code-related tasks and \citet{sorokin2022ask} in general IR, we use the same encoder $E$ for both queries and documents.
The system first encodes all documents in the database into embedding vectors and then constructs the search index. For a query $q$, it computes pairwise similarity scores between $E(q)$ and document embeddings $E(d)$ and sorts them with the dot product score
$\mathrm{score}(q, d) = E(q)^{\top} E(d)$.
Fig.~\ref{fig:model-overview} shows the system structure; we call this model \modelb (\textbf{Snippe}t \textbf{R}etrieval).

We initialize the encoder $E$ with pretrained \emph{GraphCodeBERT} (GCB)~\cite{guo_graphcodebert_2021}, a model based on RoBERTa~\cite{roberta} with 125M trainable parameters, pretrained for source code using a data flow graph along with the text representation. In our case the input is not always code, so we cannot use the data flow graph, but we discovered that even without it GCB outperforms other models.
We used the model output (last layer's hidden state) for the first $\langle s\rangle$ token as a vector representation of the input text (or code).

\textbf{Training procedure}.
The encoder is trained to maximize the similarity between a query and the matching document's embedding while minimizing the similarity between a query and embeddings of irrelevant documents. Each training sample contains one query $q$, one relevant (positive) document $d^{+}$, and $n$ irrelevant (negative) documents $D^-=\{d^{-}_{j} \}_{j=1}^n$. As the contrastive loss we use the negative log-likelihood of the positive document:

\noindent\resizebox{\linewidth}{!}{
$
\mathcal{L}(q, d^{+}, D^-) = -\log \frac{ e^{\mathrm{score}(q, d^+)}}{ \sum\limits_{j=1}^n e^{\mathrm{score}(q, d^{-}_{j})} + e^{\mathrm{score}(q, d^+)}}.$
}

\begin{figure}
\centering
\includegraphics[width=.8\linewidth]{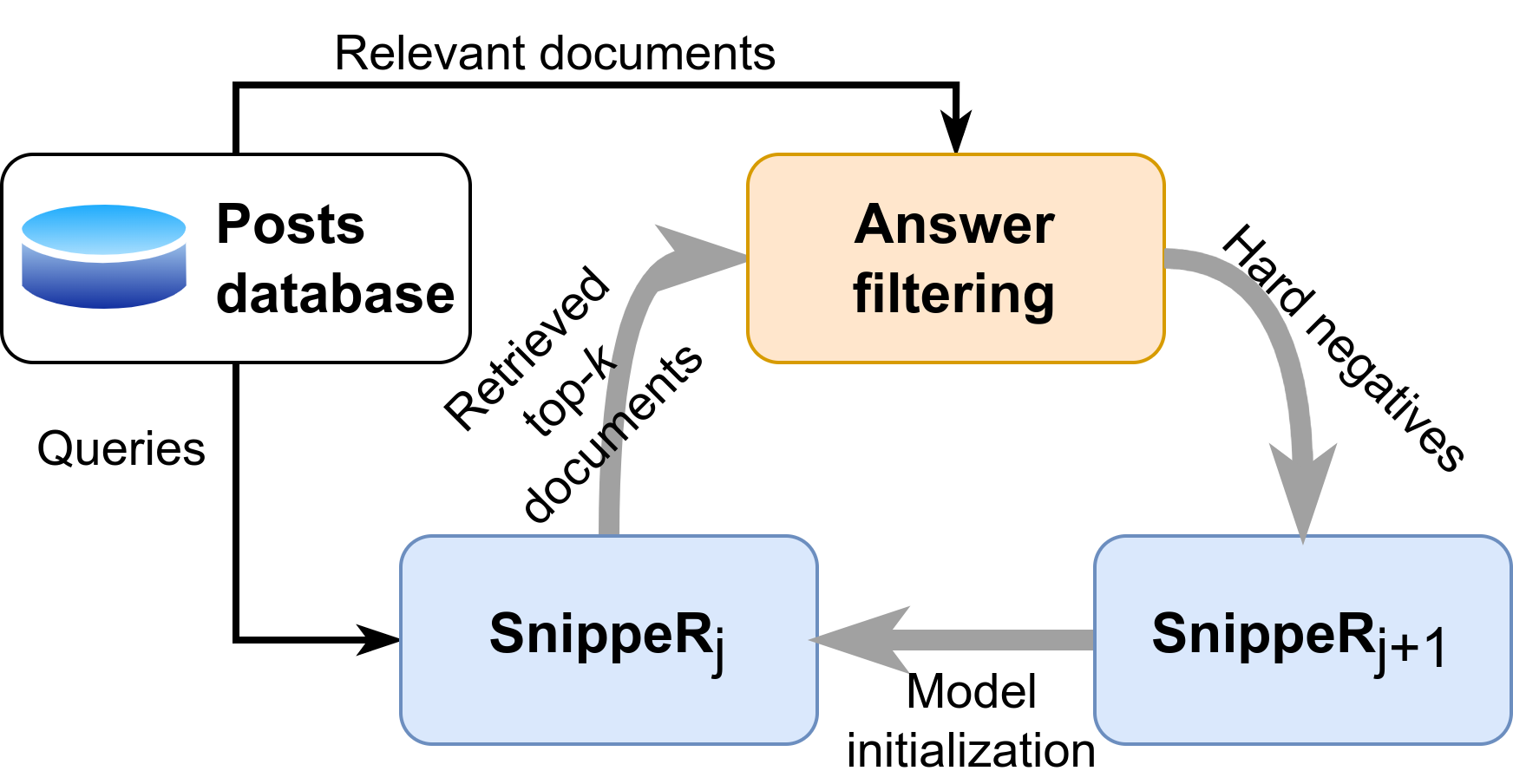}
\caption{Self-training framework.}
\label{fig:self-training}
\end{figure}

For training, we use \emph{hard negatives} mined from the previous model iteration via \emph{self-training}. We use iterative learning~\cite{qu_rocketqa_2021, izacard_distilling_2020} in the form shown in Fig.~\ref{fig:self-training}: on each step, the model retrieves top $k$ documents from the database for every training set query. Then we treat these top $k$ documents (except for the ground truth answer) as hard negative examples for the next model training iteration; in Section~\ref{sec:eval} we show the results after one such loop ($\model_2$). For each query, as other (non-hard) negatives we use other documents from the training batch and their hard negative samples; this is the~\emph{in-batch negative trick}~\cite{DPR,sorokin2022ask} that helps avoid sampling additional negatives.

\textbf{Pretraining}.
\dataset has only 1369 questions with duplicates and accepted answers in the evaluation part, but 148K pairs of duplicate questions with code snippets or tracebacks but no accepted answers. We use them in pretraining to better adapt the model for the structure and semantics of \emph{StackOverflow}. Pretraining runs the same as training with two differences:
\begin{inparaenum}[(i)]
\item for a pair of duplicates A and B we use the snippet and/or traceback from A as the query and B as the target document;
\item we include post bodies in the texts since they do not overlap with the evaluation set.
\end{inparaenum}

\textbf{Data preprocessing and training setup}.
We concatenate the code snippet $c$ and traceback $t$ (``\emph{code}'' and ``\emph{error}'' fields) to form a query $q = [c, t]$. Queries are often longer than maximum input length (256 or 512 tokens), and since the end of a traceback
usually contains crucial information such as error identifiers and meaningful error descriptions,
we remove tokens from the middle rather than the end, leaving equal number of tokens at the beginning and end.
Text documents are truncated to (the first) 256 or 512 tokens.
In \dataset, documents are represented as question title, question body, and accepted answer (``\emph{title}'', ``\emph{body}'', and ``\emph{best\_answer}'' fields).
Since queries were extracted from post bodies, we cannot use them in the ``\emph{body}'' field in the training set, so
the post body was removed from a document representation during training, leaving the model with ``\emph{title}'' and ``\emph{best\_answer}'' fields for a document. In evaluation, we use the ``\emph{body}'' field as well since now there is no issue with leaking the answer.

\section{Evaluation}\label{sec:eval}

\textbf{Setup and hyperparameters}.
We measure model performance with
$\text{Recall}@k = \sum^k_{i=1} \left[ r_i = d^+ \right]$,
where $d^+$ is the ground truth document and $r_i$ is the document retrieved at position $i$;
$k\in \{5, 10, 20, 50\}$ in our experiments.
The model was trained for 21 hours on 2 NVIDIA Tesla V100 GPUs (16GB memory each).
We used the Adam optimizer~\cite{kingma2014method} with constant learning rate schedule and 3500 warm-up steps. To stabilize training we clipped the gradient norm to~$2.0$. The learning rate was set to $10^{-5}$, batch size 12.

\begin{table}[]
\centering
\resizebox{\linewidth}{!}{\setlength{\tabcolsep}{2pt}
\begin{tabular}{l|ccccc@{}}
\toprule
& \multicolumn{4}{c}{\textbf{Recall}} \\
\textbf{Model} & \textbf{@5} & \textbf{@10} & \textbf{@20} & \textbf{@50} \\
\midrule
GraphCodeBERT~\cite{guo_graphcodebert_2021} & 0.001   & 0.001     & 0.002         & 0.009         \\
CodeBERT~\cite{codebert}    & 0.001     & 0.006     & 0.010          & 0.013                    \\
SynCoBERT~\cite{wang_syncobert_2021}   & 0.006     & 0.010      & 0.013         & 0.020                    \\
GraphCodeBERT (+CSN)   & 0.161     & 0.221      & 0.280         & 0.367                    \\
BM25~\cite{Robertson1994Okapi}        & 0.311     & 0.406          & 0.474          & 0.562      \\
\model & \textbf{0.338} & \textbf{0.451} & \textbf{0.536} & \textbf{0.657}  \\
\midrule
\end{tabular}
}
\captionof{table}{Results on \dataset.}\label{metrics}

\end{table}

\textbf{Baselines}.
We use the standard information retrieval baseline \emph{Okapi BM25}~\cite{Robertson1994Okapi}.
The dataset has a significant distribution shift between training and evaluation; since BM25 does not train, it does not suffer from the shift, which is an important factor making this baseline strong.
Other baselines include modern Transformer-based pretrained models for NLP and code understanding, trained to produce meaningful vector representations for code and/or text in the context of code search~\cite{husain_codesearchnet_2019}; e.g., CodeBERT~\cite{codebert} aims to align the embeddings of the code and corresponding text.
We also evaluated GraphCodeBERT (GCB)~\cite{guo_graphcodebert_2021} and SynCoBERT~\cite{wang_syncobert_2021} that incorporate abstract syntactic tree (AST) representations for code. ASTs are used in training but not inference since short snippets from queries often do not yield a meaningful AST.
We considered these models as base models for \model in preliminary experiments, and GCB won.
We also tried to fine-tune GraphCodeBERT on CodeSearchNet;
since our data is noisy, we fine-tuned GraphCodeBERT without ASTs
(``GraphCodeBERT (+CSN)''). All models in Table~\ref{metrics} (except BM25) are based on RoBERTa, with 125M trainable parameters.

\textbf{Results}.
Our main evaluation results are shown in Table~\ref{metrics}. Surprisingly, all Transformer-based models perform very poorly out of the box and lose to the classic BM25 baseline, despite the fact that they have been trained to embed both source code and text into a single embedding space. Fine-tuning GCB on CSN significantly improved performance, but even then GCB falls short of BM25 by a large margin.
We present the best result for \model in the table;
it has been able to outperform BM25 and all other baselines by all considered metrics. Still, the resulting Recall@5 only slightly exceeds 30\% and Recall@50 is about 65\%, which leaves significant room for improvement.

\section{Conclusion}\label{sec:conclusion}

We have presented a novel use case for code search that has not been widely studied in literature: searching \emph{by} a code snippet and/or error traceback.
We have presented a novel way to construct a dataset for this use case, leading to the \dataset dataset with about 1M queries. We have evaluated several code understanding models and found that on \dataset
they all lose even to the BM25 baseline. Thus, we have developed a new model \model for searching by code snippets and tracebacks, and have been able to outperform BM25 on \dataset. Still, absolute values of our results are relatively low, and we hope that this new setting
and dataset
will serve as a new research direction for code understanding, with \model providing a reasonable baseline.

\subsection*{Acknowledgements}
The authors are grateful to colleagues from Huawei Noah's Ark lab, especially to Irina Piontkovskaya and Wang Yasheng for organization of the collaboration which allowed this paper to happen. The work of Sergey Nikolenko and Valentin Malykh was supported by a grant for research centers in the field of artificial intelligence, provided by the Analytical Center for the Government of the Russian Federation in accordance with the subsidy agreement (agreement identifier 000000D730321P5Q0002) and the agreement with the Ivannikov Institute for System Programming of the Russian Academy of Sciences dated November 2, 2021, No. 70-2021-00142.

\section{Bibliographical References}\label{sec:reference}

\bibliographystyle{acl_natbib}
\bibliography{anthology,lit,Code-Search}

\end{document}